\documentclass{article}
\usepackage{spconf,amsmath,epsfig}
\usepackage{times}  
\usepackage{helvet} 
\usepackage{courier}  
\usepackage{amsfonts,amssymb}
\usepackage{amsmath}
\usepackage{graphicx}
\usepackage{url}
\usepackage{marvosym}

\usepackage{amssymb}
\usepackage{bm}
\let\OLDthebibliography\thebibliography
\renewcommand\thebibliography[1]{
  \OLDthebibliography{#1}
  \setlength{\parskip}{0pt}
  \setlength{\itemsep}{0pt plus 0.3ex}
}

\begin{document}\sloppy

\def\x{{\mathbf x}}
\def\L{{\cal L}}

\title{Multi-Pretext Attention Network\\for Few-shot Learning with Self-supervision}
%
\name{Hainan Li\textsuperscript{*1}\thanks{\textsuperscript{*}Equal contribution \quad \textsuperscript{\textrm{\Letter}}Corresponding author}, Renshuai Tao\textsuperscript{*1}, Jun Li\textsuperscript{1}, Haotong Qin\textsuperscript{1}, Yifu Ding\textsuperscript{1}, Shuo Wang\textsuperscript{1}, Xianglong Liu\textsuperscript{\textrm{\Letter}1}}
\address{\textsuperscript{1}State Key Lab of Software Development Environment, Beihang University\\
\{hainan,rstao,zjdyf,wangshuo2\}@buaa.edu.cn,\\
\{qinhaotong,xlliu\}@nlsde.buaa.edu.cn, junmuzi@gmail.com}
\maketitle
%

%
\begin{abstract}
Few-shot learning is an interesting and challenging study, which enables machines to learn from few samples like humans. Existing studies rarely exploit auxiliary information from large amount of unlabeled data. 
Self-supervised learning is emerged as an efficient method to utilize unlabeled data.
Existing self-supervised learning methods always rely on the combination of geometric transformations for the single sample by augmentation, while seriously neglect the endogenous correlation information among different samples that is the same important for the task.
In this work, we propose a \textbf{G}raph-driven \textbf{C}lustering (GC), a novel \textit{augmentation-free} method for self-supervised learning, which does not rely on any auxiliary sample and utilizes the endogenous correlation information among input samples. Besides, we propose \textbf{M}ulti-pretext \textbf{A}ttention \textbf{N}etwork (MAN), which exploits a specific attention mechanism to combine the traditional \textit{augmentation-relied} methods and our GC, adaptively learning their optimized weights to improve the performance and enabling the feature extractor to obtain more universal representations. We evaluate our MAN extensively on \textit{mini}ImageNet and \textit{tiered}ImageNet datasets and the results demonstrate that the proposed method outperforms the state-of-the-art (SOTA) relevant methods.~\footnote{Code can be found at \url{https://github.com/MAN-author/MAN}.}
\end{abstract}
\begin{keywords}
Few-shot learning, self-supervised learning, attention, graph convolutional network, clustering
\end{keywords}
\section{Introduction}
\label{sec:intro}
Traditional standard CNN-based models contain millions of parameters \cite{WeiOccluded2020,Qin_2020_pr,qin2019forward,qin2020bipointnet,zhang2021diversifying}, causing problems in many practical works, e.g., developing real-time interactive vision applications for portable devices, transferring knowledge from existing models to novel categories without re-training, etc. On the contrary, learning novel concepts from few examples is an interesting and challenging task with many practical advantages. In the past years, a variety of few-shot learning methods have been proposed, making great attempts to utilize the information contained in the limited labeled data.

The training data of few-shot learning is divided into base classes with a large number of labeled samples and novel classes with only a few labeled samples. Few-shot learning models acquire transferable visual analysis abilities, the typical process of which is learning representations from base classes and fine-tuning the network on novel classes to achieve the purpose of recognizing novel classes.

Most of the previous methods promote few-shot learning by expanding the amount of data of novel classes or calculating the similarity between the samples in novel classes and base classes. Optimization-based solution, as one of the most popular few-shot learning paradigms, tries to capture the relation information among the tasks, leveraging the previous learning experience as prior knowledge over tasks. Generation-based methods adopt a meta-learner for few-shot data augmentation or learn to predict classification weights for novel classes. Metric-based solution serves as another promising few-shot learning paradigm, which exploits the feature similarity information by embedding both support and query samples into a shared feature space.

Self-supervised learning is a special subcategory of unsupervised learning, which does not require manually labeled data but uses only the visual information presented in images. ``Labels'' in self-supervised learning are semi-automatically annotated during training process. The only work is to ``tell'' the network how to label the data instead of manually labeling the data. The purpose of self-supervised learning is to improve the ability of representing visual information and generate a satisfactory feature extractor, which can transfer visual analysis in other tasks. Inspired by the similarity of few-shot and self-supervised learning, some works \cite{DBLP:conf/iccv/GidarisBKPC19,su2019does} have weaved self-supervision into the training process of few-shot learning. However, existing works which integrate the two methods above usually adopt a single self-supervised method or simple fusion for multiple self-supervised methods, causing the performance to be not satisfactory.

In this work, towards improving the performance of few-shot visual learning with self-supervision, we propose a self-supervised attention network named \textbf{M}ulti-pretext \textbf{A}ttention \textbf{N}etwork (MAN), combining two types of self-supervised methods (augmentation-relied and augmentation-free) to extract the complementary information that usually neglected by one single style of self-supervised method. MAN generates an attention map to adaptively adjust the proper weights for each self-supervised method by 
endowing larger weights for ones influencing the classification objective heavily. In addition, we propose the Graph-driven Clustering method as a new clustering method for self-supervised learning by exploiting the graph convolutional network and three fully connected network layers to dynamically tune the distances between two samples. The parameters of this new clustering method are learnable, which addresses the problem of lacking back-propagation in traditional clustering methods like K-means \cite{DBLP:conf/eccv/CaronBJD18}.

The main contributions of this work are as follows:
\begin{itemize}
    \item We weave self-supervision into the training process of few-shot learning by proposing a self-supervised attention network named \textbf{M}ulti-pretext \textbf{A}ttention \textbf{N}etwork (MAN), combining augmentation-relied and augmentation-free self-supervised methods and adaptively learning the proper weights for each one by endowing larger weights for ones influencing the classification objective heavily.
    \item We propose a new clustering self-supervised method named \textbf{G}raph-driven \textbf{C}lustering (GC) in MAN by exploiting the graph convolutional network, which does not rely on auxiliary samples and utilizes the endogenous correlation information among input samples.
    \item We conduct extensive experiments on \textit{mini}ImageNet and \textit{tiered}ImageNet datasets and the results demonstrate that our method can significantly outperform a variety of the SOTA few-shot learning methods.
\end{itemize}
\section{Related work}
\noindent\textbf{Few-shot Learning.} Recently, many few-shot learning methods have been proposed. We categorize few-shot learning methods roughly into two aspects: metric learning and parameter optimization. Based on metric learning, Prototypical Networks \cite{DBLP:conf/nips/SnellSZ17} uses the average value of all samples in embedding space as the prototypical representation for each class. The classification of novel classes is regarded as the nearest neighbor problem in embedding space.
Parameter optimization methods optimize the gradient descent algorithm so that the model can quickly adapt to novel classes using only a few samples. Model-Agnostic Meta-Learning (MAML) \cite{DBLP:conf/icml/FinnAL17} is  a  classic method, however, it provides the same initialization for all tasks. To solve this problem, \cite{DBLP:conf/icml/LeeC18} learns good initialization parameters for different tasks. To improve the refining process in MAML, regularization is used in \cite{DBLP:conf/eccv/GuiWRM18} to make the gradient descent direction more accurate.


\noindent\textbf{Self-supervised learning.} Self-supervised learning uses the information of the data itself to form pseudo labels when training. And it
can be roughly divided into two branches: methods based on pretext tasks and methods based on contrastive methods.
Pretext task is used to explore the information about the data itself. In \cite{DBLP:conf/iclr/GidarisSK18}, images is rotated to $0^\circ$, $90^\circ$, $180^\circ$, $270^\circ$, and the rotation angles are predicted in pretraining process. 
\cite{DBLP:conf/iccv/DoerschGE15,DBLP:conf/eccv/NorooziF16}  predict relative position and arrangement of patches respectively. 
By constructing positive samples and negative samples, self-supervised learning based on contrastive methods considers that the distance from samples to positive samples is less than the distance to negative samples. \cite{DBLP:conf/iclr/HjelmFLGBTB19} extracts deep representational information by maximizing mutual information. To reduce calculation, \cite{DBLP:conf/cvpr/WuXYL18} proposes the concept of memory bank. MoCo \cite{DBLP:journals/corr/abs-1911-05722} solves the problem of dictionary size and consistency of memory bank. 

\begin{figure*}[t]
    \centering
    \vspace{-0.4in}
    \includegraphics[width=\textwidth]{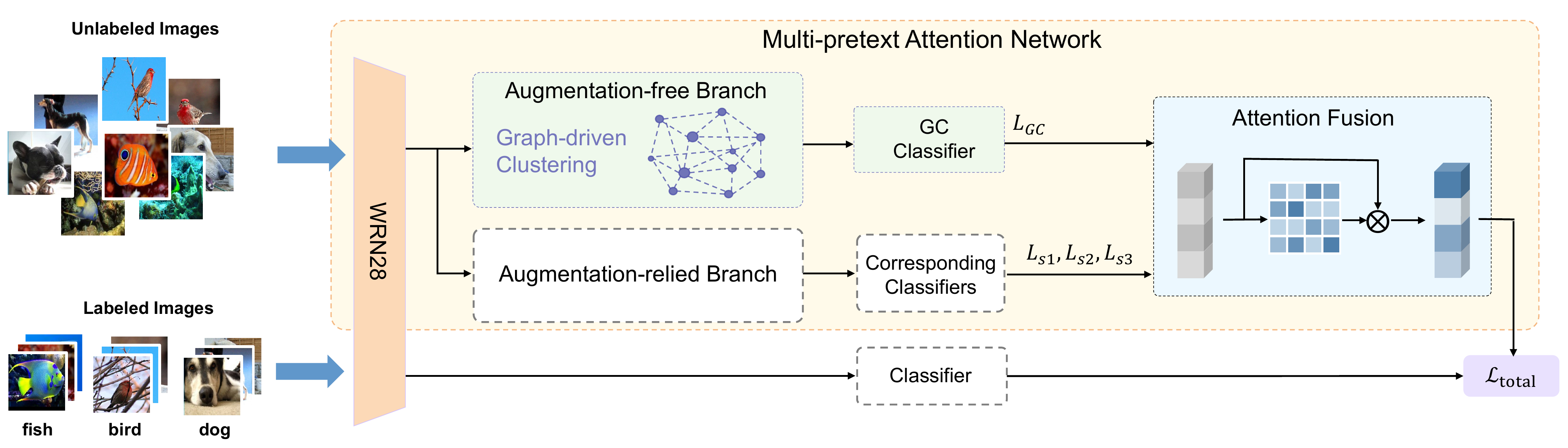} 
    \vspace{-0.3in}
    \caption{The framework for the first stage. 
    We train the feature extractor to use few-shot learning and self-supervised learning at the same time, which is implemented using a wide residual network. In the bottom path, there is a supervised classification learning. In the top path, unlabeled images are converted through two types of self-supervised methods and pseudo-labels are obtained respectively. Then, the multi-pretext attention mechanism adaptively learns the weights of the classifiers for different self-supervised methods.} 
    \label{Fig.framework} 
\end{figure*}

\section{Methods}
Common accepted practice for data augmentation is to perform a combination of geometric transformations for each single sample, which can be regarded as a kind of data augmentation and we define this type of self-supervised methods as \textit{augmentation-relied methods}. 
Our study shows that the endogenous correlation information among samples is also important to improve the model accuracy, while it is ignored by existing augmentation-relied methods.
In this work, we propose a Graph-driven Clustering method as the \textit{augmentation-free methods}, which does not generate any auxiliary samples and uses the original information of the input samples. The proposed method is able to be combined with various augmentation-relied methods (Rotation \cite{DBLP:conf/iclr/GidarisSK18}, Patches Location \cite{DBLP:conf/iccv/DoerschGE15}, Jigsaw puzzle \cite{DBLP:conf/eccv/NorooziF16}). In this section, we first introduce the preliminary of few-shot setting, and then give details to the framework and each part of our MAN.

\subsection{Preliminary}
The training data for few-shot learning is divided into two parts: base classes and novel classes. Let $D_b$ donates the training set of base classes, which contains $N_b$ different image classes $\left(\mathcal{C}_1,\mathcal{C}_2,\cdots,\mathcal{C}_{N_b}\right)$ and each class of $D_b$ has a large amount of labeled images. Also, $D_n$ donates the training set of $N_n$ classes (selected from novel classes) and each class of $D_n$ has $K$ labeled samples ($K$ = 1 or 5 in benchmarks). This setting is also called $N_n$-way $K$-shot classification. Note that the base classes and the novel classes are disjoint.

\subsection{Framework}
The framework of MAN is illustrated in Fig. \ref{Fig.framework}. In the first stage of few-shot learning, we combine supervised classification learning (the bottom path in Fig. \ref{Fig.framework}) and two types of self-supervised auxiliary methods (the top path in Fig. \ref{Fig.framework}) to train the feature extractor.
First, we input the labeled data and the unlabeled data into the Wide Residual Networks (WRN) \cite{zagoruyko2016wide} by two paths with weight sharing. Second, we preprocess the features of unlabeled images with augmentation-free method (our Graph-driven Clustering) and augmentation-relied self-supervised methods (Rotation \cite{DBLP:conf/iclr/GidarisSK18}, Patches Location \cite{DBLP:conf/iccv/DoerschGE15}, Jigsaw puzzle \cite{DBLP:conf/eccv/NorooziF16}). After extracting features by the network, the labels of the path below are used to train the classifier. And the pseudo labels of the path above are used to train the self-supervised classifiers. Here, suppose $L_{base}$ is the loss function of conventional few-shot learning method with labeled data. and $L_{s1}, L_{s2}, L_{s3}, L_{GC}$ donate the loss functions of the three self-supervised methods mentioned above and our graph-driven self-supervised clustering method, respectively. The total loss of the whole network  $L_{total}$ can be calculated by the following equation:

\begin{equation}
    L_{total} = \underset{i\sim \mathcal{I}}{\sum}\lambda_i L_{i}
\end{equation}
where $\mathcal{I} = \left\{base, s1, s2, s3, GC\right\}$ and $\lambda$ is the corresponding weight of each loss, meaning different contributions of methods to the final performance.
The proper values of all weights to make the whole network achieve the best performance will be adaptively learned by the multi-pretext attention mechanism that we introduce details in Section 3.3.

In the second stage, first, we freeze the trained parameters of feature extractor in order to keep the visual representation ability learned in the first stage. Second, we remove auxiliary models of self-supervised learning in the top path in Fig. \ref{Fig.framework} but remain the few-shot learning classifiers in the bottom path. Finally, we train the few-shot learning classifiers with samples of novel classes in $D_n$.

\subsection{Multi-pretext Attention Mechanism}
For the parameters $\lambda_{base}, \lambda_{s1}, \lambda_{s2}, \lambda_{s3}, \lambda_{GC}$, we simply set $\lambda_{base}=1.0$ because it is from traditional few-shot learning models. On the contrary, $\lambda_{s1}, \lambda_{s2}, \lambda_{s3}, \lambda_{GC}$ are learnable scalars. These parameters mean that the four self-supervised methods contribute differently to the performance of the network.
Combining multiple self-supervised methods with manually setting their weights may not achieve satisfactory performance. Besides, directly learning the weights through the deep network will cause the problem of vanishing gradient. Therefore, based on the above considerations, we propose an attention mechanism to automatically learn the weights of each self-supervised model and constrain these scalars through a sigmoid function. Further, we add 1 to each scalar outputted of the sigmoid function to keep the weights $1 < \lambda_{s1}, \lambda_{s2}, \lambda_{s3}, \lambda_{GC} < 2$ to address the problem that constraining the weight to be between 0 and 1 causes the performance to decrease. Raising the range of weight effectively reduces the loss of self-supervision, which achieves a higher accuracy.
Mathematically, the multi-pretext attention mechanism can be formulated as follows:
\begin{equation}
\mathbf{S_\lambda}=\sigma(\mathbf{L}) = \frac{1}{1+e^{-\mathbf{L}}}
\end{equation}

\begin{equation}
    L_{self} = \mathbf{L} \times \left(\mathbf{S_\lambda} + \mathbf{I}\right)\label{att}
\end{equation}
where $\mathbf{L}$ is the loss matrix consisting of four losses generated by our Graph-driven Clustering and three other traditional self-supervised models. $\mathbf{S}_\lambda$ refers to the attention map generated by our mechanism and $\mathbf{I}$ refers to the identity matrix. $L_{self}$ is the final loss of the self-supervised methods after refined we expected.

\subsection{Graph-driven Clustering}


Now we formally introduce our Graph-driven Clustering algorithm. We construct an undirected graph $\textbf{G}=\left(\textbf{V},\textbf{E}\right)$, where $\textbf{V}=\left\{\mathcal{V}_1,\dots,\mathcal{V}_{N_v}\right\}$ and $\textbf{E}=\left\{\mathcal{E}_1,\dots,\mathcal{E}_{N_e}\right\}$.
$N_v$ refers to the number of nodes in the graph and ${N_e}$ refers to the number of edges, where ${N_e}=\frac{N_v(N_v-1)}{2}$ because every two nodes have an edge connection in this graph. Note that each node represents a training sample and the value of edge between the two nodes refers to the similarity of them. We initialize the adjacency matrix $\mathbf{A}$ by calculating the cosine similarity of each two samples as follows:
\begin{equation}
    \mathbf{A}_{ij}=\mathbf{A}_{ji}=\cos\left(\mathcal{V}_i,\mathcal{V}_j\right)
\end{equation}
where $\mathbf{A}_{ij}$ refers to the value of the edge between $\mathcal{V}_i$ and $\mathcal{V}_j$.

\begin{figure}[t]
\vspace{-0.1in}
    \centering
    \includegraphics[width=0.9\columnwidth]{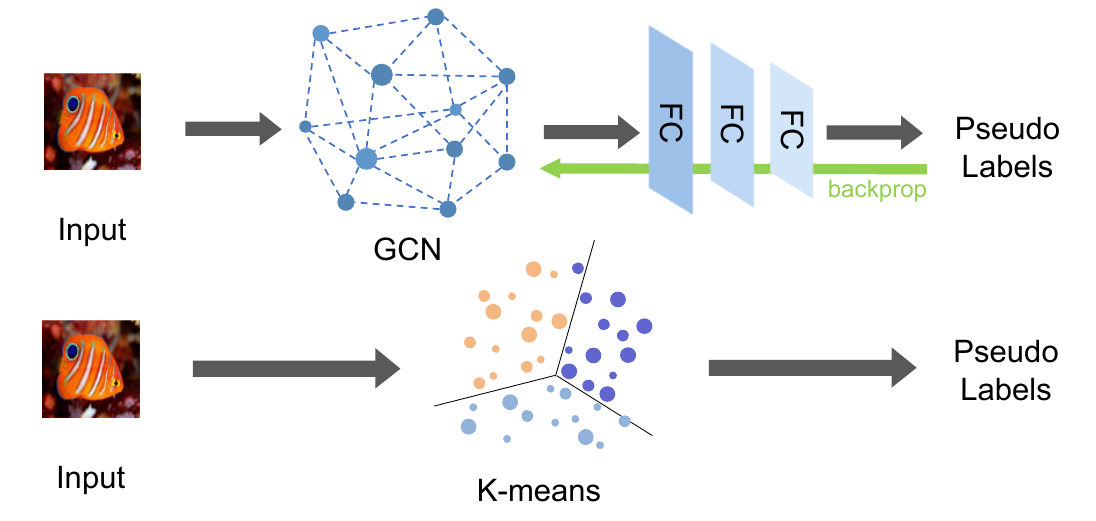}
    \vspace{-0.2in}
    \caption{Comparison of GC self-supervision and K-means clustering self-supervision} 
    \label{Fig.clustering} 
\end{figure}

Motivated by \cite{DBLP:conf/eccv/CaronBJD18}, we cluster unlabeled samples and use the result as a pseudo-label for each sample. Different from this mechanism, we exploit the graph neural network with gradient back-propagation instead of traditional clustering method like K-means without gradient back-propagation (the comparison is illustrated in Figure \ref{Fig.clustering}). 
Specifically, the feature extractor of our Graph-driven Clustering shares weights with the one in backbone network, generating feature maps of training samples. The input feature matrix, $\mathbf{X} \in \mathrm{R}^{N_v \times F \times 1 \times 1}$ where $F$ is the number of features per node, contains features of all training samples. Then we use a convolutional neural network to perform nonlinear transformations on features to improve the expression ability of the model. Finally, three fully connected layers represented as $\tau^3\left(\cdot\right)$ are applied to classify the unlabeled data.
\begin{equation}
    \mathbf{Y}_{lab} = \tau^3(\mathbf{A}\mathbf{X}\mathbf{W}+\textbf{b})
    \label{gcn}
\end{equation}
where $\mathbf{W}$ is the weight matrix and \textbf{b} is a bias matrix. Besides, $\mathbf{Y}_{lab}$ refers to the pseudo-labels generated.

Now, we input each sample $x$ and its pseudo-label $\overline{y}$ generated by our Graph-driven Clustering method to train a new classifier $C_{g}$. The loss of Graph-driven Clustering is formulated as follow:
\begin{equation}
    L_{GC}(\theta; D_u) = \underset{\left(x,\overline{y}\right)\sim D_u}{\mathbb{E}} \left[-\log C_{g}^{\overline{y}}\left(x\right)\right]\label{cluloss}
\end{equation}
where $D_u$ consists of the unlabeled samples from training set $D_b$.

\section{Experiments}

In this section, we conduct comprehensive experiments on \textit{mini}ImageNet and \textit{tiered}ImageNet datasets to evaluate the effectiveness of our MAN.

\subsection{Datasets}
\textbf{MiniImageNet} consists of 100 classes (64 for training, 16 for validation, and 20 for testing), randomly selected from ImageNet dataset \cite{DBLP:journals/ijcv/RussakovskyDSKS15}. And each class contains 600 images with size of $84 \times 84$ pixels. \textit{Tiered}ImageNet consists of 608 classes (351 for training, 97 for validation, and 160 for testing), also randomly selected from ImageNet dataset. And there are 779165 images with size of $84 \times 84$ pixels in \textit{tiered}ImageNet.

\subsection{Implementation details}

\noindent\textbf{Training in the First Stage.} 
In all experiments following, we apply the same data preprocessing on \textit{mini}ImageNet and \textit{tiered}ImageNet, which resizes the images to $80\times 80$ pixels. In order to get appropriate patches after cropping, we resize the images to $96\times 96$ pixels and then use a $3\times 3$ grid to divide the image into 9 patches with size of $32\times 32$. A harder self-supervised pretext task can help the feature extractor learn universal features, which makes the classifier more stronger and performs better in downstream tasks. Inspired by this, we apply a border jitter with patches and the patch size is set as $24\times 24$.

\noindent\textbf{Evaluation in the Second Stage.} We freeze the parameters of the feature extractor and only keep the few-shot branch ($L_{total} = L_{base}$). The second stage can be divided into two phases: few-shot training and testing. 
In the training phase, we select $N$ classes from the novel classes each time, and randomly sample $K$ samples from the selected $N$ classes to fine-tune the classifier, so as-called $N$-way $K$-shot few-shot learning. In the testing phase, we select $N\times T$ samples in the novel classes each time, where $T$ means that each class selects $T$ samples (the training samples and the test samples in the novel classes are disjoint). Note that in our experiments, we use $N = 5$, $K = 1$ and $5$ ($5$-way $1$-shot and $5$-way $5$-shot) and $T = 15$, which are standard few-shot learning settings. We evaluate our MAN through a large number of test tasks and calculate the average accuracy of all test tasks as the final results.

\begin{table}[t]
    \centering
    \setlength{\tabcolsep}{0.7mm}{
    \small
    \begin{tabular}{lccc}
        \hline
        Models&Backbone&1-shot(\%)&5-shot(\%)\\
        \hline  
        
        MAML \cite{DBLP:conf/icml/FinnAL17}&Conv-4-64&$48.70\pm 1.84$&$63.10\pm 0.92$\\
        PN \cite{DBLP:conf/nips/SnellSZ17}&Conv-4-64&$49.42\pm 0.78$&$68.20\pm 0.66$\\
        RelationNet \cite{DBLP:conf/cvpr/SungYZXTH18}&Conv-4-64&$50.40\pm 0.80$&$65.30\pm0.70$\\
        MAN(Ours)&Conv-4-64 & \textbf{51.30} $\pm$ 0.45& \textbf{68.21}  $\pm$ 0.37 \\
        \hline
        MN \cite{DBLP:conf/icml/MunkhdalaiY17}&ResNet-12&$57.10\pm 0.70$&$70.04\pm 0.63$\\
        SNAIL \cite{DBLP:conf/nips/SantoroRBMPBL17}&ResNet-12&$55.71\pm 0.99$&$68.88\pm 0.92$\\
        TADAM \cite{oreshkin2018tadam}&ResNet-12&$58.50\pm 0.30$&$76.70\pm 0.30$\\
        BF3S \cite{DBLP:conf/iccv/GidarisBKPC19}&ResNet-12&$60.76\pm 0.45$&$78.02\pm 0.34$\\
        MTL \cite{sun2019meta}&ResNet-12&$61.20\pm 1.80$&$75.50\pm 0.80$\\

        MAN(Ours)& ResNet-12 & \textbf{61.70} $\pm$ 0.47& \textbf{78.42 }$\pm$ 0.34 \\
        \hline
        CC \cite{DBLP:conf/cvpr/GidarisK18,DBLP:conf/cvpr/QiBL18}&WRN-28-10&$58.19\pm 0.45$&$76.42\pm 0.34$\\
        Qiao et al. \cite{DBLP:conf/cvpr/QiaoLSY18}&WRN-28-10&$59.60\pm 0.41$&$73.74\pm 0.19$\\
        LEO \cite{DBLP:conf/iclr/RusuRSVPOH19}&WRN-28-10&$61.76\pm 0.08$&$77.59\pm 0.12$\\
        BF3S \cite{DBLP:conf/iccv/GidarisBKPC19}&WRN-28-10&$62.93\pm 0.45$&$79.87\pm 0.33$\\
        wDAE-GNN$^\ddagger$ \cite{gidaris2019generating}&WRN-28-10&$62.96\pm 0.15$&$78.85\pm 0.10$\\

        MAN(Ours)&WRN-28-10&\textbf{64.51 }$\pm$ 0.47&\textbf{80.93 }$\pm$ 0.33\\
        \hline
    \end{tabular}
    }
    \vspace{-0.1in}
    \caption{Comparison with state-of-the-art results on miniImageNet.The average accuracies of our method are obtained by testing 2000 episodes with $95\%$ confidence intervals.The results of methods with $\ddagger$ include training set and validation set for training models.}
    \label{table1}
    \end{table}

\begin{table}[h]
\vspace{-0.1in}
\begin{center}
\setlength{\tabcolsep}{0.6mm}{
\small
\begin{tabular}{lccc}
\hline
Models&Backbone&1-shot(\%)&5-shot(\%)\\
        \hline  
        PN$^\dagger$ \cite{DBLP:conf/nips/SnellSZ17}&ResNet-12&$61.74\pm 0.77$&$80.00\pm 0.55$\\
        CTM$^\dagger$ \cite{li2019finding}&ResNet-12&$64.78\pm 0.11$&$81.05\pm 0.52$\\
        MetaOpt-SVM$^\ddagger$ \cite{lee2019meta}&ResNet-12&$65.81\pm 0.74$&$81.75\pm 0.53$\\
        MAN(Ours)&ResNet-12& \textbf{65.99} $\pm$ 0.51&\textbf{81.97} $\pm$ 0.37\\
        \hline
        LEO$^\ddagger$ \cite{DBLP:conf/iclr/RusuRSVPOH19}&WRN-28-10&$66.33\pm 0.05$&$81.44\pm 0.09$\\
        wDAE-GNN$^\ddagger$ \cite{gidaris2019generating}&WRN-28-10&$68.18\pm 0.16$&$83.10\pm 0.12$\\

        MAN(Ours)&WRN-28-10&\textbf{70.35} $\pm$ 0.51&\textbf{84.90} $\pm$ 0.36\\
        \hline
    \end{tabular}
    }
    \end{center}
    \vspace{-0.2in}
    \caption{Comparison with state-of-the-art 5-way results on tieredImageNet.The average accuracies of our method are obtained by testing 2000 episodes with $95\%$ confidence intervals. The results of methods with $\dagger$ are from \cite{simon2020adaptive}. The results of methods with $\ddagger$ are from \cite{simon2020adaptive} and include training set and validation set for training models.}
    \label{table2}
    \end{table}
\subsection{Comparison with prior work}
We compare various previous methods on different backbone networks, the traditional convolutional network \cite{DBLP:conf/nips/VinyalsBLKW16}, ResNet \cite{lee2019meta} and Wide Residual Network \cite{zagoruyko2016wide} on both \textit{mini}ImageNet and \textit{tiered}ImageNet. Tab. \ref{table1} shows the effectiveness of our MAN on \textit{mini}ImageNet in the cases of different backbone networks.
According to our experiments, in previous works, RelationNet achieves the best performance in both 1-shot and 5-shot settings in the case of traditional convolutional backbone network, MTL and wDAE-GNN achieving the best performance in 1-shot setting in other two backbone networks, BF3S achieving the best performance in 5-shot setting in other two backbone networks. As Tab. \ref{table1} illustrates, the performance of our MAN is much better than traditional few-shot methods in all of three backbone network. Our method outperforms RelationNet by $0.90\%$ in 1-shot setting and 2.99\% in 5-shot setting. Besides, our method outperforms MTL and BF3S by $0.50\%$ and 2.92\% in 1-shot setting and 5-shot setting, respectively, in the case of ResNet backbone network. In the case of WRN backbone network, our method outperforms wDAE-GNN and BF3S by 1.55\% and 1.08\% in 1-shot setting and 5-shot setting, respectively.

Tab. \ref{table2} reports the results on \textit{tiered}ImageNet. As Tab. \ref{table2} illustrates, compared with the most advanced few-shot learning wDAE-GNN method, we only use the training set to train the model and freeze the backbone network during verification, which is more difficult. However, our method outperforms the most advanced wDAE-GNN method by 2.17\% in 5-way 1-shot learning and 1.80\% in 5-way 5-shot learning. The results of Tab. \ref{table2} are obtained by testing 2000 episodes with $95\%$ confidence intervals, which is a common setting in few-shot learning methods.

To verify the effectiveness of the proposed Graph-driven Clustering method, we replace it with the traditional K-means clustering method and carry out the same training and evaluation process. As Tab. \ref{table4} shows, the Graph-driven Clustering method outperforms the K-means clustering method by 1.75\% in the case of they are implemented into the baseline individually without other self-supervised methods. Besides, the Graph-driven Clustering method outperforms the K-means clustering method by 0.56\% in the case of other supervised methods and the attention mechanism are implemented.

\begin{table}[t]
\vspace{-0.1in}
    \centering
    \resizebox{.85\columnwidth}{!}
    {
    \begin{tabular}{lcc}
        \hline
        Methods&1-shot(\%)&5-shot(\%)\\
        \hline
        CC \cite{DBLP:conf/cvpr/GidarisK18,DBLP:conf/cvpr/QiBL18}&$58.19\pm 0.45$&$76.42\pm 0.34$\\
        K-means&$58.66\pm 0.45$&$76.62\pm 0.35$\\
        GC&$60.41\pm 0.46$&$77.46\pm 0.34$\\
        \hline
        MAN (K-means)&$63.95\pm 0.46$&$80.10\pm 0.34$\\
        MAN (GC)&\textbf{64.51} $\pm$ 0.47& \textbf{80.93} $\pm$ 0.33\\
        \hline
    \end{tabular}
    }
    \vspace{-0.1in}
    \caption{The comparison results between our Graph-driven and K-means clustering methods with the WRN-28-10 backbone network on miniImageNet dataset.}
    \label{table4}
    \end{table}


\subsection{Ablation study}


In this section, we study the effects of different parts in our MAN on \textit{mini}ImageNet and the results are reported in Tab. \ref{table5}.


\textbf{Effect of Graph-driven Clustering Method.} In order to explore the impact of our Graph-driven Clustering method, we implement the network without the Graph-driven Clustering module. We use the self-supervised branch to conduct all the comparative experiments. In the experiments of rotation prediction and patches location self-supervised methods, we simply use the models provided in \cite{DBLP:conf/iccv/GidarisBKPC19}. These results demonstrate that (1) self-supervised methods can improve few-shot learning and a combination of multiple self-supervision can further promote the improvement, and (2) our Graph-driven Clustering method is effective, which improves the accuracy in both 1-shot and 5-shot settings by 0.52\% and 0.99\%. 

\textbf{Effect of Multi-pretext Attention Mechanism.} To evaluate how multi-pretext attention mechanism affects the performance, we replace the attention mechanism by a simple summation of loss functions of the four self-supervised methods. As Tab. \ref{table5} demonstrates, the multi-pretext attention mechanism further improves 0.66\% and 0.71\% in 1-shot and 5-shot settings, compared with direct summation.

\begin{table}[t]
\vspace{-0.1in}
    \centering
    \resizebox{.99\columnwidth}{!}
    {
    \begin{tabular}{lcc}
        \hline
        Methods&1-shot(\%)&5-shot(\%)\\
        \hline
        CC \cite{DBLP:conf/cvpr/GidarisK18,DBLP:conf/cvpr/QiBL18}&$58.19\pm 0.45$&$76.42\pm 0.34$\\
        Rot&$62.81\pm 0.46$&$80.01\pm 0.34$\\
        Loc&$60.70\pm 0.47$&$77.61\pm 0.33$\\
        Jig&$61.16\pm 0.47$&$77.86\pm 0.35$\\
        GC &$60.41\pm 0.46$&$77.46\pm 0.34$\\
        \hline
        MAN (Rot+Loc+Jig) &$63.99\pm 0.47$&$79.94\pm 0.35$\\
        SUM (Rot+Loc+Jig+GC)&$63.85\pm 0.46$&$80.22\pm 0.34$\\
        MAN (Ours)&\textbf{64.51} $\pm$ 0.47&\textbf{80.93} $\pm$ 0.33\\
        \hline
    \end{tabular}
    }
    \vspace{-0.1in}
    \caption{The results of ablation study with the WRN-28-10 back-bone network on miniImageNet dataset.}
    \label{table5}
    \vspace{-0.1in}
    \end{table}
\vspace{-0.1in}
\section{Conclusion}
To exploit auxiliary information from large amount of unlabeled data for few-shot learning, in this work, we propose a \textbf{G}raph-driven \textbf{C}lustering (GC), a novel \textit{augmentation-free} method for self-supervised learning, which does not rely on any auxiliary sample and utilizes the endogenous correlation information among input samples. Besides, we propose \textbf{M}ulti-pretext \textbf{A}ttention \textbf{N}etwork (MAN), which exploits a specific attention mechanism to combine the traditional \textit{augmentation-relied} methods and our GC, adaptively learning their optimized weights to improve the performance and enabling the feature extractor to obtain more universal representations. We evaluate our MAN extensively on \textit{mini}ImageNet and \textit{tiered}ImageNet datasets and the results demonstrate that the proposed method outperforms the state-of-the-art (SOTA) relevant methods. We hope that our contributions can promote the rapid development of few-shot learning with self-supervision.

\section{Acknowledgement}
This work was supported by The National Key Research and Development Plan of China (2020AAA0103502), National Natural Science Foundation of China (62022009, 61872021), Beijing Nova Program of Science and Technology (Z191100001119050), and State Key Lab of Software Development Environment (SKLSDE-2020ZX-06).

\small
\bibliographystyle{IEEEbib}
\bibliography{icme2021template}

\end{document}